\pdfoutput=1

\documentclass[11pt]{article}

\usepackage[]{EMNLP2023}

\usepackage{times}
\usepackage{latexsym}
\usepackage{amsmath}
\usepackage[T1]{fontenc}

\usepackage[utf8]{inputenc}

\usepackage{microtype}

\usepackage{inconsolata}

\usepackage{braket}
\usepackage{verbatim}
\usepackage{ulem}
\usepackage{adjustbox}
\usepackage{multirow}
\usepackage{enumitem}

\title{UNLEARN\\Efficient Removal of Knowledge in Large Language Models}

\author{ Tyler Lizzo, Larry Heck\\ AI Virtual Assistant (AVA) Lab\\
   Georgia Institute of Technology  \\ \texttt{\{lizzo,larryheck\}@gatech.edu}}

\begin{document}
\maketitle

\begin{abstract}
Given the prevalence of large language models (LLMs) and the prohibitive cost of training these models from scratch, dynamically forgetting specific knowledge e.g., private or proprietary, without retraining the model has become an important capability.  This paper proposes a novel method to achieve this objective called $\mbox{UNLEARN}$. The approach builds upon subspace methods to identify and specifically target the removal of knowledge without adversely affecting other knowledge in the LLM. Results demonstrate 96\% of targeted knowledge can be forgotten while maintaining performance on other knowledge within 2.5\% of the original model, significantly outperforming the discriminatory abilities of the previous state-of-the-art. A dual method called LEARN is also proposed for targeted knowledge addition. Results show LEARN can match the fine-tuning accuracy of Low-Rank Adaptation (LoRA) without adversely affecting similar tasks. 
\end{abstract}

\section{Introduction}
The swift advancement and widespread deployment of large language models (LLMs) have brought many challenges including the inability to remove knowledge from the LLMs at will. Efficient removal of knowledge has become increasingly important with `Right to be Forgotten' laws \cite{goldman2020introduction} and Europe's \textit{General Data Protection Regulation} \cite{goddard2017gdpr}. Traditional training methodologies often lack the flexibility and efficiency required to address both tasks, especially when rapid model adaptation is needed without comprehensive retraining.

This paper introduces UNLEARN, a novel algorithm that can forget or unlearn knowledge within an LLM without adversely affecting related knowledge. UNLEARN leverages subspace techniques to identify the subspaces spanned by particular knowledge (tasks) and discrimination methods to separate that subspace from subspaces of similar tasks. This allows the algorithm to prevent performance degradation when there are similar tasks, a common issue with traditional methods and of particular importance to data privacy regulations. Further, this technique uses a unified set of operators, where the task matrices are identical and used to either enhance or reduce the model's performance for a given task. 

UNLEARN achieves 96\% forgetting on the task of interest while maintaining performance on dissimilar tasks within 2.5\% of the original model. When the tasks are similar, UNLEARN still achieves nearly 80\% forgetting on the task of interest while preserving performance on similar tasks within 10\%.  These results significantly outperform the state-of-the-art, which achieves similar forgetting but is accompanied by significant degradation on similar tasks.

The forgetting of UNLEARN can easily be converted to {\it add knowledge} to the LLM.  This new method LEARN matches the fine-tuning accuracy of the LoRA method \cite{hu2021lora} {\em without affecting related tasks}, demonstrating its dual nature across both knowledge unlearning and fine-tuning scenarios.

The contributions of this work are as follows:
\begin{itemize}[itemsep=1mm, parsep=0pt]
\item An efficient method to identify the subspace of specific knowledge within an LLM.
\item A novel approach called subspace discrimination and task removal to selectively target and remove specific knowledge without adversely affecting other knowledge in the LLM.
\item The introduction of LEARN, a dual algorithm to UNLEARN that provides a new approach to adding new knowledge to the LLM without affecting its other knowledge.
\end{itemize}
This paper presents the UNLEARN algorithm and demonstrates its performance in removing knowledge represented as tasks. Section \ref{relatedworks} reviews the literature on Parameter Efficient Fine-Tuning, Machine Unlearning, and LLM Unlearning. Section \ref{methods} describes the three main parts of UNLEARN: subspace identification, subspace discrimination, and task removal. In Section \ref{experiments}, the performance of UNLEARN is tested over a large set of metrics and settings and compared to the current state-of-the-art. Section \ref{LEARN} introduces LEARN, a dual application of the UNLEARN algorithm for adding knowledge to the LLM. A comparison to traditional fine-tuning methods is made in Section \ref{LoRA}. Future works are discussed in Section \ref{futureworks}. Finally, Section \ref{conclusion} concludes the paper and outlines potential directions for future research.

\section{Related Works}\label{relatedworks}
\subsection{Parameter Efficient Fine-Tuning}
Parameter Efficient Fine-Tuning (PEFT) is used to fine-tune large models without modifying most of the original pre-trained weights, resulting in significant computational and storage savings.

One of the most significant PEFT methods is Low-Rank Adaptation \citep[LoRA;][]{hu2021lora}, which decomposes weight updates into two low-rank matrices. While reducing trainable parameters by 10,000 times and GPU memory usage by 3 times, LoRA is still able to maintain the fine-tuning performance of a systems. Quantized Low-Rank Adapation would build upon LoRA's performance gains by quantizing model weights \citep{dettmers2023qlora}. 

Other notable PEFT methods include prompt tuning \citep{lester2021power,qin2021learning}, tuning hidden states \citep[IA$^3$;][]{liu2022fewshot}, adding layers \cite{houlsby2019parameterefficient}, tuning the embedding layer inputs \citep{an2022inputtuning}, and hybrid approaches \citep{mahabadi2021compacter}. These extend prior work on domain adaptation of deep neural networks for Natural Language Processing \cite{jaech2016domain}. 

\subsection{Machine Unlearning}
Machine unlearning is the process of removing the influence of data on an already trained model, creating a model that behaves as if it was never trained on that data \citep{xu2023machine}. Its origins are in data protection regulations, such as the \textit{California Consumer Privacy Act} \citep[CCPA;][]{goldman2020introduction} and the European Union's \textit{General Data Protection Regulation} \citep[GDPR;][]{goddard2017gdpr}, which assert a user's `Right to be Forgotten,' the right to have their personal data erased upon request.

Machine unlearning has since been extended to myriad areas:  federated learning \citep{liu2022,zhang2023}, image classification \citep{bourtoule2021,gupta2021,liu2024}, and image generation \citep{gandikota2023erasing,kumari2023ablating,fan2024salun}.

The most rigorous method for machine unlearning is `exact' unlearning, completely retraining a model with the data points of interest removed \citep{yan2022arcane,nguyen2022survey,fan2024salun}. Although exact unlearning guarantees the removal of data, it is impractical for models of any significant size due to the high computation cost. For instance, training Llama 2 70B took $\sim$ 1.7 million GPU-hours on Nvidia A100 GPUs \citep{touvron2023llama2openfoundation}.

\subsection{LLM Unlearning}\label{llmunlearning}
There is an increasing interest in machine unlearning in the context of LLMs \citep{jang2022knowledge,meng2023locating,liu2024safer}. Important works have demonstrated the need for machine unlearning within LLMs, showing clear motivations from both regulatory and application-specific standpoints \citep{zhang2023right,liu2024rethinking}. 

Existing methods for LLM unlearning include gradient ascent to reascend the learning curve \citep{jang2022knowledge,chen2023unlearn,yao2024llmunlearning}, preference optimization using alternative responses \citep{eldan2023whos,maini2024tofu}, and input-based approaches \citep{pawelczyk2024incontext,thaker2024guardrail}. 

However, these methods face significant challenges. There are the aforementioned cost and time restraints. The vast amounts of training data used for LLM training adds to the complexity, as identifying and isolating the specific data points to be unlearned is a non-trivial task \citep{eldan2023whos,ilharco2023editing}. The scope of unlearning is generally underspecified; unlearning should remove knowledge within the scope of the targeted data while maintaining performance on other data \citep{mitchell2022memorybased}. Finally, there is a lack of comprehensive evaluation methods to assess the effectiveness of unlearning in LLMs \citep{patil2023sensitive,shi2024detecting}.

\section{UNLEARN Method} \label{methods}
\begin{figure*}[t]
    \centering
    \includegraphics[width=\linewidth]{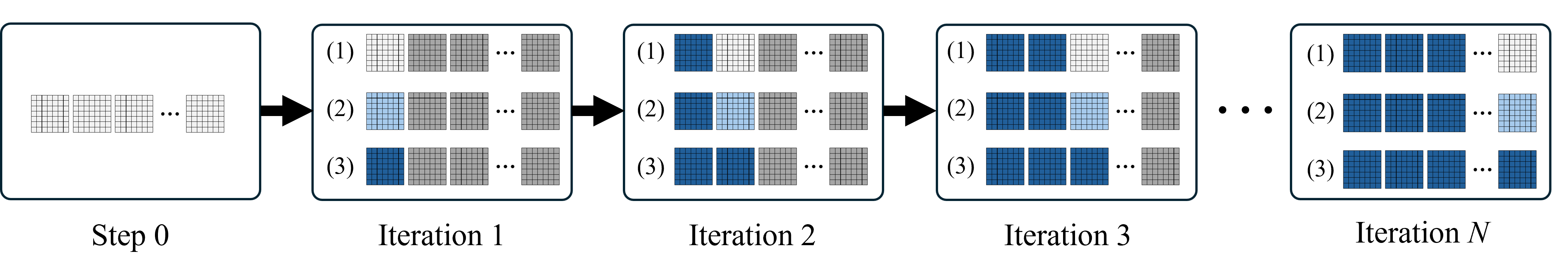}
    \caption{The Subspace Identification Process. The process begins by randomly initializing the model weights and then freezing them. Then an iterative process of unfreezing, training, and refreezing each layer occurs. This results in a set of matrices that capture an accurate representation for that task.}
    \label{fig:identification}
\end{figure*}

The method proposed in this paper consists of three main tasks: subspace identification, discrimination, and removal. Subspace identification trains a knowledge (task)-dependent matrix for a specified layer while freezing all other layers. This sequential, layer-by-layer training starts with the first layer and progresses through the entire network to yield a set of matrices that represent the task-dependent subspace (Section \ref{identification}). Once identified, subspace discrimination removes the information unique to the task of interest while preventing any degradation of other tasks. This is achieved using a variation of the Gram-Schmidt process to orthogonalize subspaces, allowing mutual information to be preserved (Section \ref{discrimination}). The final step is subspace removal, where the modified task matrix, $T_i'$, is subtracted (Section \ref{removal}).

\subsection{Subspace Identification} \label{identification}
This step identifies the subspace of a specific task within the LLM weight space. 
The method utilizes a general training that is implemented layer-by-layer, starting with the first layer ($l=1)$. All training is performed with a train/validation/test split of 0.6/0.2/0.2: The train set is used for training the network, the validation set determines when to stop training for a specific layer in our sequential process, and all evaluations are performed on the final test set:

\begin{enumerate}
\setcounter{enumi}{-1}
    \item \textbf{Model:} The original pretrained weights of the LLM are removed and the weights for all layers are randomly initialized.
    \item \textbf{Layer Freezing:} Except for the weights at layer $l$, all other weights for the subsequent layers of an LLM are frozen to isolate the training to one layer at a time.
    \item \textbf{Training:} Training is completed on the task dataset with the $l$-th layer unfrozen. This is achieved by maximizing the conditional language modeling objective:
\begin{equation}
\max_{T_i^l} \sum_{(x,y)\in {\cal Z}} \sum_{t=1}^{|y|} \log (P_{T_i^l}(y_t | x,y_{<t}))
\end{equation}
where $x_i$ and $y_i$ are sequences of tokens and 
$T_i^l \in \Re^{n\times n}$ is the matrix for task $i$ at the $l$-th layer and $n\times n$ the dimensions of the original pre-trained weight matrix. 

Given the matrix $T_i^l$ is trained on a specific task, the matrix is likely rank deficient. To facilitate training, we alter each layer using a bottleneck architecture as shown in Figure \ref{fig:bottleneck} with interior dimension $k$, where $T_i^l = F G$.  
\begin{figure}[h]
    \centering
    \includegraphics[width=0.3\textwidth]{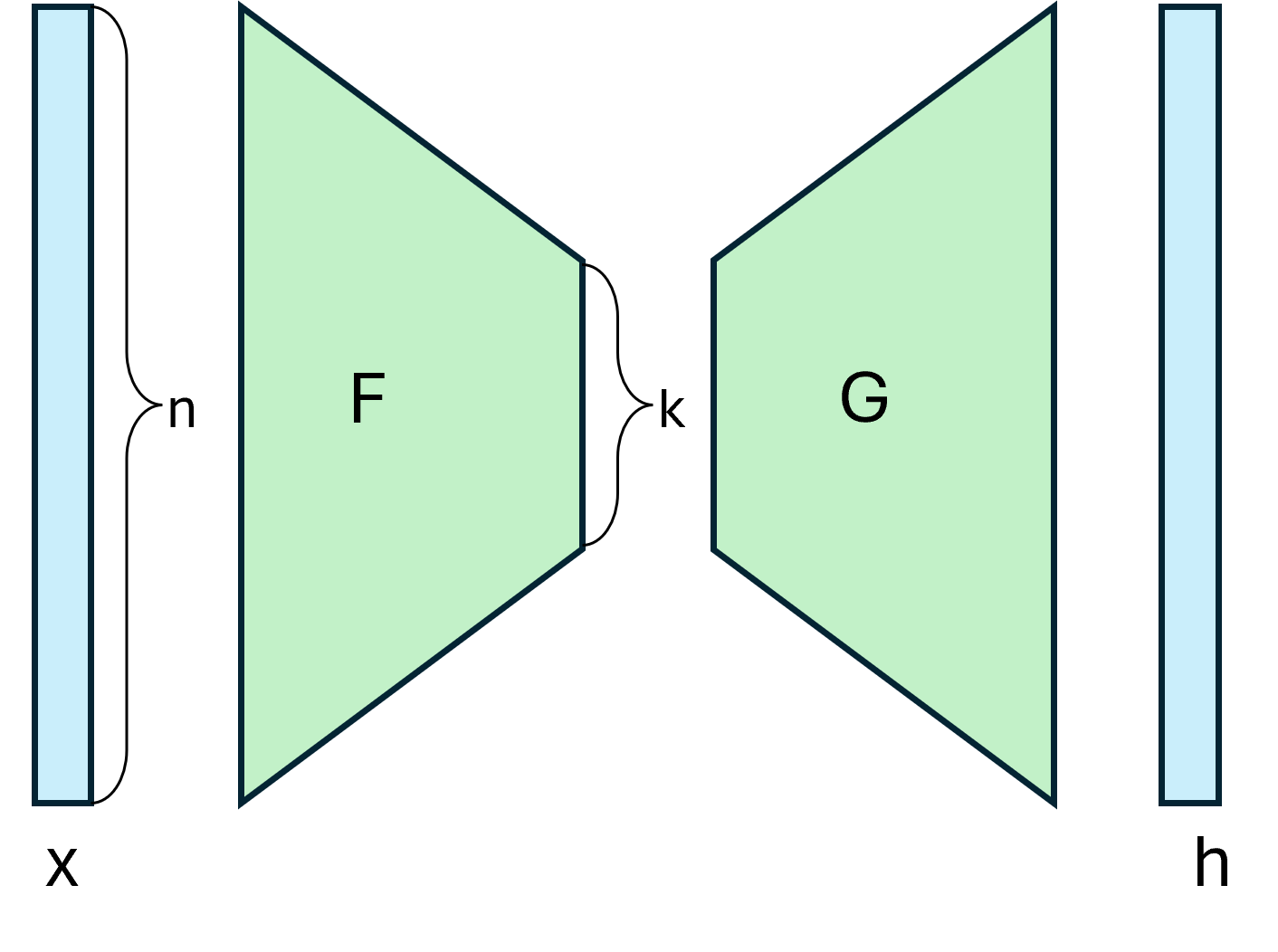}
    \caption{Bottleneck architecture of layer $l$ with interior dimension $k \ll n$}
    \label{fig:bottleneck}
\end{figure}
    
    \item \textbf{Sequential Training:} Once the training at layer $l$ is complete, that layer is frozen and the next layer is unfrozen. For our experiments, training concluded once loss on the validation set had stopped decreasing (i.e. potential overfitting of the training set was starting). Similar training is then performed on the next layer. This process is repeated across all layers, resulting in weight matrices for each layer.
\end{enumerate}

By the end of this sequential training and freezing process, shown in Figure \ref{fig:identification}, the set of weight matrices captures an accurate representation of the task-dependent subspace within the weights of the Transformer model. This method is lightweight, maintaining the computational efficiency of low rank training. The layer-by-layer approach was taken because the early layers contain higher-level semantic information, while the later layers contain more task/fact-specific information. Training in this method ensures the most reliable identification of the tasks.
\subsection{Subspace Discrimination} \label{discrimination}
Once a task-dependent subspace has been identified, it could be removed by subtracting it from the entire weight space (layer-by-layer). While this may be effective at removing the task of interest, it leads to performance loss when similar tasks are also evaluated, i.e. ones that occupy similar subspaces. Therefore, a method is required that maintains the mutual information between these two subspaces, only removing the information unique to the task of interest. We call this {\em subspace discrimination}.

To achieve subspace discrimination, we utilize a variation of the Gram-Schmidt process. Gram-Schmidt is used to orthogonalize a set of vectors in an inner product space. Given the subspace $U$ spanned by vectors $u_1,\cdots,u_N$, we can find the orthogonal subspace to a vector $v_k$ with the following:
\[
v'_k=v_k-\sum_{j=1}^{N}\frac{\braket{v_k,u_j}}{\braket{u_j,u_j}}u_j.
\]
A proof that $v_k'$ is orthogonal to all $u_j$ is offered in Appendix \ref{proof_orthogonality}. For our application, we compute:
\begin{footnotesize}
\begin{align*}
SV_k(T'_i)=SV_k(T_i)-\sum_{j=1}^N\frac{SV_k(T_i)\cdot SV_j(T_o)}{SV_j(T_o)\cdot SV_j(T_o)}SV_j(T_o)
\end{align*}
\end{footnotesize}
where
$T_i$ represents the identified subspace to be removed, $T_o$ represents a similar task, and  
$SV_k(T_i')$ represents the $k$-th singular vector of matrix $T_i$ for one of the Transformer layers $l$. When applied to two tasks, every pair of weight matrices is decomposed and separated in this manner. For three or more tasks, the other task matrices; $T_{o,1}$, $T_{o,2}$, $\cdots$, $T_{o,n}$, are added into one $T_o$ matrix, then the above equation is applied. We chose to use Euclidean inner products, inspired by the original LORA paper \cite{hu2021lora}, which demonstrated that efficient training could be achieved with linear rank decompositions. While neural network parameter spaces are non-Euclidean, the practical success of the LORA method justified our approach.

Initially, the similarity of tasks was determined subjectively. However, this subspace discrimination method allows us to quantify task similarity, as there will be more overlap in the weight space of two similar sets of matrices. For two dissimilar tasks, the discrimination process will have no effect, as they are already orthogonal.

Subspace discrimination is essential to the $\mbox{UNLEARN}$ algorithm, allowing for the precise separation of task-specific information within shared weight spaces and ensuring that the removal of one task does not undesirably impact the performance on similar tasks. Consequently, subspace discrimination enhances the algorithm's adaptability and robustness.

\subsection{Task Removal} \label{removal}
The final step removes the task subspace. To achieve this, our approach uses SVD reconstruction to reconstitute the modified task matrix, $T_i'$ from the singular values of $T_i$ and singular vectors $SV(T_i')$ above.  Once $T_i'$ is computed, we subtract it from $W'$ for each matrix in the LLM:
\[\widehat{W'}=W'-T_i'\]

\section{Experiments}\label{experiments}
All experiments in this section use the same setup, with Llama 2 70b serving as the LLM. For the training step in the subspace identification method (Section \ref{identification}) as illustrated in Figure \ref{fig:bottleneck}, we used the Python package \textsc{loralib} \cite{hu2021lora} but, rather than training a fine-tuning adapter, we modified it to train the bottleneck in Figure \ref{fig:bottleneck} from scratch. We used a rank of $k=16$. Only the attention matrices were modified during training. This was inspired by the original LORA paper \cite{hu2021lora}, where they only adapted the attention weights.


\subsection{Datasets}
A diverse selection of benchmarks is essential to evaluate performance degradation across similar tasks when modifying task-specific subspaces within LLMs. This study used two signification collections of benchmarks: Holistic Evaluation of Language Models \citep[HELM;][]{liang2023holistic} and the Beyond-the-Imitation-Game Benchmark \citep[BIG-Bench;][]{srivastava2023imitation}.

HELM evaluates a wide range of use cases and metrics, encompassing general language abilities to simple question-answering settings. This benchmark evaluates models across multiple metrics--accuracy, fairness, robustness, efficiency, and more--providing a detailed view into the general language capabilities of models.

Complementing HELM, BIG-Bench focuses on more specific and niche tasks that probe the boundaries of current LLM capabilities. With 204 tasks contributed from experts across fields, BIG-Bench was invaluable for testing specific tasks that were beyond the domain of HELM. Importantly, BIG-Bench provided niche tasks that have little overlap with other tasks, offering an unbiased perspective on subspace removal.

Together, these datasets facilitate a comprehensive analysis of the influence of subspace removal on LLM performance across a spectrum of tasks. By integrating the thorough evaluation of HELM for general language abilities with the specialized tasks from BIG-Bench, this study explores how manipulation of tasks affects both broad and targeted model capabilities. This sheds light on the ability of $\mbox{UNLEARN}$ to remove a task without affecting adjacent tasks.

\subsection{Single Task Removal} \label{exp:str}
\begin{table*}[t]
\centering
\caption{Performance of $\mbox{UNLEARN}$ on a variety of tasks, compared to three state-of-the-art models: Gradient Ascent  (Yao et al., 2024), Knowledge Gap Alignment (KGA, Wang et al., 2023), and Knowledge Unlearning (KU, Jang et al., 2022). Targeted Task represents the task that was 'unlearned'. The tasks of interest are NarrativeQA (NQA), NaturalQuestions (NQ), Massive Multitask Language Understanding (MMLU), IMDB benchmark for sentiment analysis in movies (IMDB), Real-world Annotated Few-Shot (RAFT), Grade School Math 8K (GSM8K), and arithmetic. The green columns represent the targeted task and the yellow columns represent the similar task.}
\label{table:unlearn}
\begin{tabular}{|l|l|ccccccc|}
\hline
\multicolumn{1}{|l|}{}              & Model         & \multicolumn{7}{c|}{Evaluation Tasks}                      \\ \cline{3-9} 
\multicolumn{1}{|l|}{}              &               & NQA   & NQ    & MMLU  & IMDB  & RAFT  & GSM8K & arithmetic \\ \cline{2-9} 
\multicolumn{1}{|l|}{Targeted Task} & Base Model    & 0.778 & 0.680  & 0.583 & 0.952 & 0.719 & 0.483 & 0.991      \\ \hline
\multirow{5}{*}{GSM8K}              & Gradient Ascent & 0.768 & 0.651 & 0.574 & 0.949 & 0.710 & \cellcolor{green!25}0.052 & \cellcolor{yellow!25}0.574      \\
                                    & KGA           & 0.767 & 0.664 & 0.561 & 0.937 & 0.718 & \cellcolor{green!25}0.136 & \cellcolor{yellow!25}0.682      \\
                                    & KU            & 0.763 & 0.666 & 0.574 & 0.933 & 0.716 & \cellcolor{green!25}0.043 & \cellcolor{yellow!25}0.487      \\
                                    & UNLEARN w/o D & 0.758 & 0.681 & 0.577 & 0.949 & 0.715 & \cellcolor{green!25}0.017 & \cellcolor{yellow!25}0.633      \\
                                    & UNLEARN       & 0.772 & 0.674 & 0.582 & 0.946 & 0.723 & \cellcolor{green!25}0.087 & \cellcolor{yellow!25}0.956      \\ \hline
\multirow{5}{*}{arithmetic}         & Gradient Ascent & 0.782 & 0.663 & 0.577 & 0.953 & 0.713 & \cellcolor{yellow!25}0.215 & \cellcolor{green!25}0.084      \\
                                    & KGA           & 0.767 & 0.675 & 0.581 & 0.939 & 0.700 & \cellcolor{yellow!25}0.105 &\cellcolor{green!25} 0.017      \\
                                    & KU            & 0.760 & 0.672 & 0.567 & 0.942 & 0.719 & \cellcolor{yellow!25}0.183 & \cellcolor{green!25}0.063      \\
                                    & UNLEARN w/o D & 0.757 & 0.680 & 0.578 & 0.949 & 0.716 &\cellcolor{yellow!25} 0.087 & \cellcolor{green!25}0.028      \\
                                    & UNLEARN       & 0.771 & 0.681 & 0.569 & 0.955 & 0.712 & \cellcolor{yellow!25}0.461 & \cellcolor{green!25}0.825      \\ \hline
\multirow{5}{*}{NQA}                & Gradient Ascent & \cellcolor{green!25}0.094 & \cellcolor{yellow!25}0.415 & 0.573 & 0.945 & 0.709 & 0.469 & 0.978      \\
                                    & KGA           & \cellcolor{green!25}0.183 & \cellcolor{yellow!25}0.229 & 0.581 & 0.942 & 0.717 & 0.482 & 0.976      \\
                                    & KU            & \cellcolor{green!25}0.163 &\cellcolor{yellow!25} 0.329 & 0.569 & 0.949 & 0.701 & 0.479 & 0.976      \\
                                    & UNLEARN w/o D & \cellcolor{green!25}0.118 &\cellcolor{yellow!25} 0.263 & 0.567 & 0.966 & 0.702 & 0.466 & 0.976      \\
                                    & UNLEARN       & \cellcolor{green!25}0.135 &\cellcolor{yellow!25} 0.628 & 0.581 & 0.969 & 0.723 & 0.460 & 0.989      \\ \hline
\multirow{5}{*}{NQ}                 & Gradient Ascent & \cellcolor{yellow!25}0.483 & \cellcolor{green!25}0.184 & 0.554 & 0.94 & 0.693 & 0.477 & 0.963      \\
                                    & KGA           & \cellcolor{yellow!25}0.501 &\cellcolor{green!25} 0.243 & 0.557 & 0.946 & 0.697 & 0.479 & 0.989      \\
                                    & KU            &\cellcolor{yellow!25} 0.416 & \cellcolor{green!25}0.113 & 0.558 & 0.926 & 0.712 & 0.468 & 0.973      \\
                                    & UNLEARN w/o D &\cellcolor{yellow!25} 0.419 & \cellcolor{green!25}0.142 & 0.570 & 0.936 & 0.717 & 0.464 & 0.979      \\
                                    & UNLEARN       &\cellcolor{yellow!25} 0.703 &\cellcolor{green!25} 0.147 & 0.567 & 0.941 & 0.716 & 0.471 & 0.983      \\ \hline
\end{tabular}
\end{table*}

The first trio of experiments evaluated the $\mbox{UNLEARN}$ method using only subspace identification (Section \ref{identification}) and task removal (Section \ref{removal}) without the subspace discrimination method (Section \ref{discrimination}). In these experiments, a single task was removed and performance across a set of tasks was observed. We will refer to these experiments as $\mbox{`UNLEARN w/o D'}$, where `w/o D' refers to the absence of subspace discrimination.

In the first experiment, the math word problem dataset GSM8K \citep{cobbe2021training} was removed using the $\mbox{`UNLEARN w/o D'}$ method. This is the first Targeted Task in Table \ref{table:unlearn}. The first six columns under Evaluation Tasks were chosen because they are very different tasks from GSM8K, ranging from question-answering \cite[NarrativeQA;][]{kocisky2017narrativeqa} to more general benchmarks \cite[MMLU;][]{hendrycks2021measuring}. Because these tasks are dissimilar, they theoretically have little overlap in their weight subspaces.
Evaluating the six chosen benchmarks on both the base model and $\mbox{'UNLEARN w/o D'}$ model shows our approach successfully forgets (dropped performance) by 96.5\% on the desired GSM8K task, while all other tasks had minimal degradation (less than 2.5\%).

For the second experiment, an additional benchmark was added, the arithmetic benchmark from BIG-Bench. The base model performs exceptionally well on this task, as do most off-the-shelf LLMs; however, it was important for demonstrating what happens when there are two similar tasks, as arithmetic is quite similar to GSM8K. Again, the $\mbox{'UNLEARN w/o D'}$ algorithm was applied to GSM8K. This time, the GSM8K benchmark was not the only metric affected; the arithmetic metric was also affected (down 33\%), shown in Table \ref{table:unlearn}.

The final experiment was similar to the second; however, the arithmetic benchmark was used to train our $T_i$ matrices. Table \ref{table:unlearn} shows that the arithmetic benchmark performance degrades by 97\% while the GSM8K metric worsens by 82\%. Comparing these last two experiments, the removal of a simpler task leads to greater degradation of the more complex task compared to the reverse.

This outcome underscores the challenges with task-specific subspace removal when dealing with closely aligned tasks. The performance decline on the second task suggests that the extracted subspace on the first task contains features shared by the second's subspace, highlighting the need for the subspace discrimination technique of Section \ref{discrimination}.

\subsection{Task Discrimination}
Prompted by the shared degradation seen on arithmetic and GSM8K in Section \ref{exp:str}, these experiments explore the connection between closely related tasks and evaluate the efficacy of the subspace discrimination method proposed in Section \ref{discrimination}. These experiments aimed to orthongonally separate the subspaces corresponding to two tasks, allowing us to manipulate one subspace while preserving the integrity of the other.  
These experiments focus on two sets of overlapping tasks: NarrativeQA/NaturalQuestions and arithmetic/GSM8K.

The first pair of tasks both involve question answering: NarrativeQA \citep{kocisky2017narrativeqa} answers questions over books or movie scripts, while NaturalQuestions \citep{kwiatkowski} answers questions from Google search queries. Two separate experiments were run: one with NarrativeQA as the task of interest and one with NaturalQuestions as the task of interest. As seen in Table \ref{table:unlearn}, when NarrativeQA is the task of interest, $\mbox{UNLEARN}$ successfully reduces its performance while the performance on NaturalQuestions is relatively unaffected (down 7.5\%). Similarly, when NaturalQuestions is the task of interest, NarrativeQA's performance is mostly preserved while NaturalQuestions's performance is successfully reduced.

The second pair of tasks is arithmetic and GSM8K. When the subspace discrimination method is applied to GSM8K, performance on GSM8K successfully decreases while performance on arithmetic is preserved. However, when the subspace discrimination method is applied to arithmetic as the task of interest, there is no degradation in the performance of either metric. This behavior can be explained by the relative simplicity of the arithmetic benchmark; its subspace is likely encapsulated within the subspace of the GSM8K metric.


\subsection{Optimal Rank}
We explore the impact of varying the rank of the rank-deficient matrices during subspace identification, as shown in Figure \ref{fig:bottleneck}. For the NaturalQuestions vs NarrativeQA experiment of the $\mbox{UNLEARN}$ approach, the rank was varied: $\mbox{$k=$1,2,4,8,16,32}$. We found that the performance is not hindered for $k$ values above 4 as seen in $\mbox{Table \ref{table:rank}}$. However, there is a slight degradation of performance on the tasks of interest for the lower-rank experiments; the task of interest was not forgotten as effectively, and the similar task experienced greater performance degradation. This result can be attributed to the subspace identification step not capturing the subspaces for those tasks as accurately when the rank is lower. 

These results suggest that the rank can be significantly reduced with minimal performance loss. This is reasonable given that the subspaces of interest were quite small compared to the overall dimensions of the weight matrices. We hypothesize that the minimum rank required for full performance would vary slightly with the complexity of the task. These insights provide a valuable direction for optimizing the efficiency of the $\mbox{UNLEARN}$ method, especially in resource-constrained environments.
\begin{table*}[h]
\centering
\caption{Performance of $\mbox{UNLEARN}$ when the rank (k) is modified. Targeted Task represents the task that was `unlearned'. The tasks of interest are NarrativeQA (NQA), NaturalQuestions (NQ), Massive Multitask Language Understanding (MMLU), IMDB benchmark for sentiment analysis in movies (IMDB), Real-world Annotated Few-Shot (RAFT), Grade School Math 8K (GSM8K), and arithmetic.}
\label{table:rank}
\begin{tabular}{|c|c|ccccccc|}
\hline
k                   & Targeted Task & \multicolumn{7}{c|}{Evaluation Tasks}                      \\ \hline
                    &               & NQA   & NQ    & MMLU  & IMDB  & RAFT  & GSM8K & arithmetic \\ \cline{3-9} 
Base Model          &               & 0.778 & 0.68  & 0.583 & 0.952 & 0.719 & 0.483 & 0.991      \\ \hline
\multirow{2}{*}{1}  & NQA           & 0.167 & 0.599 & 0.58  & 0.938 & 0.702 & 0.464 & 0.974      \\
                    & NQ            & 0.684 & 0.198 & 0.582 & 0.951 & 0.701 & 0.482 & 0.989      \\ \hline
\multirow{2}{*}{2}  & NQA           & 0.151 & 0.609 & 0.564 & 0.931 & 0.712 & 0.479 & 0.987      \\
                    & NQ            & 0.688 & 0.173 & 0.58  & 0.95  & 0.701 & 0.466 & 0.97       \\ \hline
\multirow{2}{*}{4}  & NQA           & 0.128 & 0.624 & 0.568 & 0.946 & 0.703 & 0.471 & 0.971      \\
                    & NQ            & 0.711 & 0.152 & 0.567 & 0.934 & 0.718 & 0.482 & 0.986      \\ \hline
\multirow{2}{*}{8}  & NQA           & 0.136 & 0.627 & 0.58  & 0.931 & 0.718 & 0.475 & 0.99       \\
                    & NQ            & 0.701 & 0.152 & 0.579 & 0.931 & 0.698 & 0.468 & 0.974      \\ \hline
\multirow{2}{*}{16} & NQA           & 0.135 & 0.628 & 0.581 & 0.969 & 0.723 & 0.46  & 0.989      \\
                    & NQ            & 0.703 & 0.147 & 0.567 & 0.941 & 0.716 & 0.471 & 0.983      \\ \hline
\multirow{2}{*}{32} & NQA           & 0.134 & 0.619 & 0.579 & 0.937 & 0.704 & 0.467 & 0.974      \\
                    & NQ            & 0.704 & 0.156 & 0.583 & 0.933 & 0.696 & 0.483 & 0.98       \\ \hline
\end{tabular}
\end{table*}
\subsection{Using $\mbox{UNLEARN}$ to LEARN}\label{LEARN}
\subsubsection{LEARN methodology}
The $\mbox{UNLEARN}$ methodology, initially designed for the selective removal of task-specific information from LLMs, also presents a versatile framework that can be adapted for the enhancement of model performance on particular tasks. This section explores `LEARN,' the application of our earlier $\mbox{UNLEARN}$ algorithm for training on new information. This method aims to {\it add knowledge} and/or amplify the representation of a given task within the model, leading to improved performance on that task.

The LEARN approach uses the same principles as $\mbox{UNLEARN}$ but inverts the application to focus on task enhancement. Specifically, the method involves identifying the subspace associated with a desired task using the approach in Section \ref{identification}; this step is identical to $\mbox{UNLEARN}$. The difference comes with task addition instead of task removal; the only necessary change is flipping the equation for task removal from Section \ref{removal}:
\[W'=W+T_i'\]
This addition should bolster performance on a new task, as the $T_i'$ sits on top of the existing weight matrix, similar to the function of most LLM adapters. In addtion, due to subspace discrimination (Section \ref{discrimination}), adding the new knowledge should have minimal adverse effects on other knowledge already in the LLM.

\subsubsection{LEARN evaluation}
To evaluate the effect of the LEARN method, experiments were conducted on tasks where pre-trained models showed suboptimal performance but had the potential to perform well if fine-tuned. Identifying tasks that meets these criteria for larger LLMs (50 B+ parameter) is challenging because they are trained on such extensive datasets that it is more difficult to find data not included in the training set. Therefore, by restricting the size of the LLM, we limit the total learning capacity of the model, allowing us to squeeze out additional learning that the LLM should be able to handle. 

These experiments used a similar setting to before, with the exception of using Llama 2 7b. The dataset of interest is LegalBench, a benchmark built by a collaboration between lawyers and ML engineers to measure legal reasoning in LLMs \citep{guha2023legalbench}. Llama 2 7b performs  between 30-50\% across all tasks, leaving room for improvement.

When the LEARN algorithm was applied to the model for LegalBench, it showed marked improvement across all tasks. Table \ref{table:learn} shows the consistent improvement across tasks and a 40\% boost to the average performance of the system compared to the base LLM.
\begin{table}[t]
\centering
\caption{Performance of LEARN and LoRA on LegalBench}
\label{table:learn}
\begin{tabular}{lccc}
\hline
Task           & Base Model & LEARN & LoRA \\ \hline
Issue          & 50.1       &  73.4   & 72.9    \\
Rule           & 42.7       &  61.8   & 63.1    \\
Conclusion     & 53.9       &  69.3   & 69.6    \\
Interpretation & 48.1       &  68.1   & 67.4    \\
Rhetorical     & 45.4       &  62.5   & 61.2    \\ \hline
Average        & 48.0       &  67.0   & 66.8    \\ \hline
\end{tabular}
\end{table}
Training with LEARN is shown relative to traditional LoRA fine-tuning. Only the two tasks of interest were shown in Table \ref{table:learn} because there was a similar lack of impact on the other tasks. LEARN matches the performance of LoRA. By systematically adding task-specific subspaces, LEARN fine-tunes the model's performance on a selected task and minimizes any degradation of other capabilities due to the subspace discrimination method. The dual capability of $\mbox{UNLEARN}$/LEARN underscores its main value: the ability to use the same training runs for both forgetting and learning.

\section{Comparison to Existing Methods}\label{LoRA}
This section presents a comparative analysis of the $\mbox{UNLEARN}$/LEARN methodology against existing methods, with a focus on generality and task performance.

\subsection{Generality and Efficiency}
A key advantage of $\mbox{UNLEARN}$/LEARN is its operational flexibility. It offers a generalized framework that can be applied to full fine-tuning or any PEFT method for fine-tuning. $\mbox{UNLEARN}$/LEARN applies the same underlying principles in any setting--either adding or subtracting task-specific matrices from the model's weight matrices--to both enhance (LEARN) and diminish ($\mbox{UNLEARN}$) the model's performance on specific tasks. Because the same set of matrices are being used regardless of algorithm, this simplifies model management and reduces the computational and storage overhead. 

\subsection{Task Performance}
In scenarios involving similar tasks, the differences between $\mbox{UNLEARN}$/LEARN and existing methods become even more pronounced. In the LEARN setting of Table \ref{table:learn}, both methods show comparable improvements in task performance, demonstrating their efficacy for bolstering model performance. In the forgetting setting, the $\mbox{UNLEARN}$ algorithm is able to successfully discriminate between two similar tasks and only remove the task of interest. 

We compared $\mbox{UNLEARN}$ to the current state-of-the-art algorithms: Gradient Ascent \cite{yao2024llmunlearning}, Knowledge Gap Alignment \cite[KGA;][]{wang2023kga}, and Knowledge Unlearning \cite[KU][]{jang2022knowledge}. As seen in Table \ref{table:unlearn}, these state-of-the-art methods are unable to discriminate effectively between tasks, leading to performance degradation in closely related tasks. For example, when NarrativeQA is the task of interest, $\mbox{UNLEARN}$ successfully degrades that task (down from 0.778 to 0.135) while maintaing the performance on NaturalQuestions (from 0.680 to 0.628). All three state-of-the-art algorithms successfully degrade NarrativeQA: GA degrades the task to 0.094, KGA to 0.183, and KU to 0.163. However, they all show significantly diminished performance on NaturalQuestions: GA degrades the task to 0.415, KGA to 0.229, and KU to 0.329. These state-of-the-art methods lack the discrimination ability to target the knowledge they seek to remove without unwanted performance effects on secondary tasks.

Conversely, with its precise subspace manipulation, the $\mbox{UNLEARN}$ method allows for the selective removal of task influences without negatively impacting the performance of related tasks. This specificity is particularly beneficial in multi-task learning/unlearning environments where tasks share overlapping features (similar weight subspaces). As such, $\mbox{UNLEARN}$ is better suited for forgetting tasks while preserving similar tasks.

\section{Future Works}\label{futureworks}
This paper has laid the groundwork for several intriguing avenues for future research. First, while our initial work focused on removing broad domain knowledge, future efforts will extend this methodology to the removal of specific knowledge and facts. We are currently collecting datasets that will facilitate this extension, particularly in scenarios involving private or harmful information.

There are some scalability concerns if UNLEARN is applied to a large number of tasks. While the current work targets the selective removal of a small number of unwanted tasks, future research will investigate strategies to efficiently handle discrimination between larger sets of similar tasks.

Our current approach was largely inspired by the original LORA paper \cite{hu2021lora}, which was our motivation for only manipulating the attention weights. Subsequent research into LORA revealed the effectiveness of manipulating the other layers within an LLM. Future works will explore the adaption of other layers to enhance the flexibility and performance of UNLEARN.

\section{Conclusion}\label{conclusion}
This paper introduces $\mbox{UNLEARN}$, a novel approach for forgetting selected knowledge in Large Language Models. This method relies on subspace identification for tasks and subspace discrimination between similar tasks. The experimental results demonstrate significant performance gains, highlighting the effect of $\mbox{UNLEARN}$ on removing unwanted knowledge without having deleterious effects on related tasks. The method's ability to isolate and remove specific subspaces within the model ensures precise unlearning, making it a valuable tool for managing the complexities of task forgetting. 

Compared to state-of-the-art methods like Gradient Ascent, $\mbox{UNLEARN}$ offers substantial advantages in terms of generality, efficiency, and precision. $\mbox{UNLEARN}$ achieves 96\% forgetting on the task of interest while maintaining performance on other tasks within 2.5\% of the original model. When similar tasks are considered, $\mbox{UNLEARN}$ achieves nearly 80\% forgetting on the task of interest while preserving performance on the similar task within 10\% of the original model. The discriminative ability of $\mbox{UNLEARN}$ far outpaces that of the existing state-of-the-art, ensuring targeted unlearning without compromising the performance on related tasks.

\section*{Limitations}
Although $\mbox{UNLEARN}$ enhances the abilities of LLMs to forget knowledge, certain limitations still need to be addressed. One limitation is when tasks completely overlap, as observed with arithmetic and GSM8K. When a subspace is entirely contained within another, as arithmetic was within GSM8K, it becomes challenging to discriminate between these two tasks. This highlights the distinction between knowledge and the metrics that measure knowledge, which we will explore this distinction in future works.

Another limitation of this paper that will be addressed in future work is to more fully leverage the experimental insights to optimize the efficiency of the UNLEARN method. 

\section*{Ethics Statement}
While $\mbox{UNLEARN}$ has significant potential benefits, such as improving model flexibility and efficiency, we are also mindful of the ethical implications. By allowing models to forget specific tasks, we enhance privacy and security by ensuring that sensitive information can be effectively removed. This is particularly important in contexts where models are trained on private or confidential data. Further, $\mbox{UNLEARN}$ can promote fairness by removing biased or harmful information.

However, there is also a risk that such methods could be misused to intentionally modify important information, leading to biased outputs. We advocate for the transparent and responsible use of this technology, with appropriate safeguards and policies to prevent such misuse.

\section*{Acknowledgements}
This work was supported by CoCoSys, one of seven centers in JUMP 2.0, a Semiconductor Research Corporation (SRC) program sponsored by DARPA.

\bibliography{emnlp2023}

\begin{thebibliography}{46}
\expandafter\ifx\csname natexlab\endcsname\relax\def\natexlab#1{#1}\fi

\bibitem[{aet al.(2019)}]{kwiatkowski}
Tom~Kwiatkowski aet al. 2019.
\newblock Natural questions: a benchmark for question answering research.
\newblock \emph{Transactions of the Association of Computational Linguistics}.

\bibitem[{An et~al.(2022)An, Li, Lin, Liu, Chen, Fu, Chen, Zheng, and
  Lou}]{an2022inputtuning}
Shengnan An, Yifei Li, Zeqi Lin, Qian Liu, Bei Chen, Qiang Fu, Weizhu Chen,
  Nanning Zheng, and Jian-Guang Lou. 2022.
\newblock \href {http://arxiv.org/abs/2203.03131} {Input-tuning: Adapting
  unfamiliar inputs to frozen pretrained models}.

\bibitem[{Bourtoule et~al.(2021)Bourtoule, Chandrasekaran, Choquette-Choo, Jia,
  Travers, Zhang, Lie, and Papernot}]{bourtoule2021}
Lucas Bourtoule, Varun Chandrasekaran, Christopher~A. Choquette-Choo, Hengrui
  Jia, Adelin Travers, Baiwu Zhang, David Lie, and Nicolas Papernot. 2021.
\newblock \href {https://doi.org/10.1109/SP40001.2021.00019} {Machine
  unlearning}.
\newblock In \emph{2021 IEEE Symposium on Security and Privacy (SP)}, pages
  141--159.

\bibitem[{Chen and Yang(2023)}]{chen2023unlearn}
Jiaao Chen and Diyi Yang. 2023.
\newblock \href {http://arxiv.org/abs/2310.20150} {Unlearn what you want to
  forget: Efficient unlearning for llms}.

\bibitem[{Cobbe et~al.(2021)Cobbe, Kosaraju, Bavarian, Chen, Jun, Kaiser,
  Plappert, Tworek, Hilton, Nakano, Hesse, and Schulman}]{cobbe2021training}
Karl Cobbe, Vineet Kosaraju, Mohammad Bavarian, Mark Chen, Heewoo Jun, Lukasz
  Kaiser, Matthias Plappert, Jerry Tworek, Jacob Hilton, Reiichiro Nakano,
  Christopher Hesse, and John Schulman. 2021.
\newblock \href {http://arxiv.org/abs/2110.14168} {Training verifiers to solve
  math word problems}.

\bibitem[{Dettmers et~al.(2023)Dettmers, Pagnoni, Holtzman, and
  Zettlemoyer}]{dettmers2023qlora}
Tim Dettmers, Artidoro Pagnoni, Ari Holtzman, and Luke Zettlemoyer. 2023.
\newblock \href {http://arxiv.org/abs/2305.14314} {Qlora: Efficient finetuning
  of quantized llms}.

\bibitem[{Eldan and Russinovich(2023)}]{eldan2023whos}
Ronen Eldan and Mark Russinovich. 2023.
\newblock \href {http://arxiv.org/abs/2310.02238} {Who's harry potter?
  approximate unlearning in llms}.

\bibitem[{et~al.(2023{\natexlab{a}})}]{srivastava2023imitation}
Aarohi~Srivastava et~al. 2023{\natexlab{a}}.
\newblock \href {http://arxiv.org/abs/2206.04615} {Beyond the imitation game:
  Quantifying and extrapolating the capabilities of language models}.

\bibitem[{et~al.(2023{\natexlab{b}})}]{guha2023legalbench}
Neel~Guha et~al. 2023{\natexlab{b}}.
\newblock \href {http://arxiv.org/abs/2308.11462} {Legalbench: A
  collaboratively built benchmark for measuring legal reasoning in large
  language models}.

\bibitem[{et~al.(2023{\natexlab{c}})}]{liang2023holistic}
Percy~Liang et~al. 2023{\natexlab{c}}.
\newblock \href {http://arxiv.org/abs/2211.09110} {Holistic evaluation of
  language models}.

\bibitem[{Fan et~al.(2024)Fan, Liu, Zhang, Wong, Wei, and Liu}]{fan2024salun}
Chongyu Fan, Jiancheng Liu, Yihua Zhang, Eric Wong, Dennis Wei, and Sijia Liu.
  2024.
\newblock \href {http://arxiv.org/abs/2310.12508} {Salun: Empowering machine
  unlearning via gradient-based weight saliency in both image classification
  and generation}.

\bibitem[{Gandikota et~al.(2023)Gandikota, Materzynska, Fiotto-Kaufman, and
  Bau}]{gandikota2023erasing}
Rohit Gandikota, Joanna Materzynska, Jaden Fiotto-Kaufman, and David Bau. 2023.
\newblock \href {http://arxiv.org/abs/2303.07345} {Erasing concepts from
  diffusion models}.

\bibitem[{Goddard(2017)}]{goddard2017gdpr}
Michelle Goddard. 2017.
\newblock \href {https://doi.org/10.2501/IJMR-2017-050} {The eu general data
  protection regulation (gdpr): European regulation that has a global impact}.
\newblock \emph{International Journal of Market Research}, 59(6):703--705.

\bibitem[{Goldman(2020)}]{goldman2020introduction}
Eric Goldman. 2020.
\newblock An introduction to the california consumer privacy act (ccpa).
\newblock \emph{Santa Clara Univ. Legal Studies Research Paper}.

\bibitem[{Gupta et~al.(2021)Gupta, Jung, Neel, Roth, Sharifi-Malvajerdi, and
  Waites}]{gupta2021}
Varun Gupta, Christopher Jung, Seth Neel, Aaron Roth, Saeed Sharifi-Malvajerdi,
  and Chris Waites. 2021.
\newblock \href
  {https://proceedings.neurips.cc/paper_files/paper/2021/file/87f7ee4fdb57bdfd52179947211b7ebb-Paper.pdf}
  {Adaptive machine unlearning}.
\newblock In \emph{Advances in Neural Information Processing Systems},
  volume~34, pages 16319--16330. Curran Associates, Inc.

\bibitem[{Hendrycks et~al.(2021)Hendrycks, Burns, Basart, Zou, Mazeika, Song,
  and Steinhardt}]{hendrycks2021measuring}
Dan Hendrycks, Collin Burns, Steven Basart, Andy Zou, Mantas Mazeika, Dawn
  Song, and Jacob Steinhardt. 2021.
\newblock \href {http://arxiv.org/abs/2009.03300} {Measuring massive multitask
  language understanding}.

\bibitem[{Houlsby et~al.(2019)Houlsby, Giurgiu, Jastrzebski, Morrone,
  de~Laroussilhe, Gesmundo, Attariyan, and
  Gelly}]{houlsby2019parameterefficient}
Neil Houlsby, Andrei Giurgiu, Stanislaw Jastrzebski, Bruna Morrone, Quentin
  de~Laroussilhe, Andrea Gesmundo, Mona Attariyan, and Sylvain Gelly. 2019.
\newblock \href {http://arxiv.org/abs/1902.00751} {Parameter-efficient transfer
  learning for nlp}.

\bibitem[{Hu et~al.(2021)Hu, Shen, Wallis, Allen-Zhu, Li, Wang, Wang, and
  Chen}]{hu2021lora}
Edward~J. Hu, Yelong Shen, Phillip Wallis, Zeyuan Allen-Zhu, Yuanzhi Li, Shean
  Wang, Lu~Wang, and Weizhu Chen. 2021.
\newblock \href {http://arxiv.org/abs/2106.09685} {Lora: Low-rank adaptation of
  large language models}.

\bibitem[{Ilharco et~al.(2023)Ilharco, Ribeiro, Wortsman, Gururangan, Schmidt,
  Hajishirzi, and Farhadi}]{ilharco2023editing}
Gabriel Ilharco, Marco~Tulio Ribeiro, Mitchell Wortsman, Suchin Gururangan,
  Ludwig Schmidt, Hannaneh Hajishirzi, and Ali Farhadi. 2023.
\newblock \href {http://arxiv.org/abs/2212.04089} {Editing models with task
  arithmetic}.

\bibitem[{Jaech et~al.(2016)Jaech, Heck, and Ostendorf}]{jaech2016domain}
Aaron Jaech, Larry Heck, and Mari Ostendorf. 2016.
\newblock Domain adaptation of recurrent neural networks for natural language
  understanding.
\newblock \emph{arXiv preprint arXiv:1604.00117}.

\bibitem[{Jang et~al.(2022)Jang, Yoon, Yang, Cha, Lee, Logeswaran, and
  Seo}]{jang2022knowledge}
Joel Jang, Dongkeun Yoon, Sohee Yang, Sungmin Cha, Moontae Lee, Lajanugen
  Logeswaran, and Minjoon Seo. 2022.
\newblock \href {http://arxiv.org/abs/2210.01504} {Knowledge unlearning for
  mitigating privacy risks in language models}.

\bibitem[{Kocisky et~al.(2017)Kocisky, Schwarz, Blunsom, Dyer, Hermann, Melis,
  and Grefenstette}]{kocisky2017narrativeqa}
Tomas Kocisky, Jonathan Schwarz, Phil Blunsom, Chris Dyer, Karl~Moritz Hermann,
  Gábor Melis, and Edward Grefenstette. 2017.
\newblock \href {http://arxiv.org/abs/1712.07040} {The narrativeqa reading
  comprehension challenge}.

\bibitem[{Kumari et~al.(2023)Kumari, Zhang, Wang, Shechtman, Zhang, and
  Zhu}]{kumari2023ablating}
Nupur Kumari, Bingliang Zhang, Sheng-Yu Wang, Eli Shechtman, Richard Zhang, and
  Jun-Yan Zhu. 2023.
\newblock \href {http://arxiv.org/abs/2303.13516} {Ablating concepts in
  text-to-image diffusion models}.

\bibitem[{Lester et~al.(2021)Lester, Al-Rfou, and Constant}]{lester2021power}
Brian Lester, Rami Al-Rfou, and Noah Constant. 2021.
\newblock \href {http://arxiv.org/abs/2104.08691} {The power of scale for
  parameter-efficient prompt tuning}.

\bibitem[{Liu et~al.(2022{\natexlab{a}})Liu, Tam, Muqeeth, Mohta, Huang,
  Bansal, and Raffel}]{liu2022fewshot}
Haokun Liu, Derek Tam, Mohammed Muqeeth, Jay Mohta, Tenghao Huang, Mohit
  Bansal, and Colin Raffel. 2022{\natexlab{a}}.
\newblock \href {http://arxiv.org/abs/2205.05638} {Few-shot parameter-efficient
  fine-tuning is better and cheaper than in-context learning}.

\bibitem[{Liu et~al.(2024{\natexlab{a}})Liu, Luo, and Zhu}]{liu2024}
Mengda Liu, Guibo Luo, and Yuesheng Zhu. 2024{\natexlab{a}}.
\newblock Machine unlearning with affine hyperplane shifting and maintaining
  for image classification.
\newblock In \emph{Neural Information Processing}, pages 215--227, Singapore.
  Springer Nature Singapore.

\bibitem[{Liu et~al.(2024{\natexlab{b}})Liu, Yao, Jia, Casper, Baracaldo, Hase,
  Xu, Yao, Li, Varshney, Bansal, Koyejo, and Liu}]{liu2024rethinking}
Sijia Liu, Yuanshun Yao, Jinghan Jia, Stephen Casper, Nathalie Baracaldo, Peter
  Hase, Xiaojun Xu, Yuguang Yao, Hang Li, Kush~R. Varshney, Mohit Bansal, Sanmi
  Koyejo, and Yang Liu. 2024{\natexlab{b}}.
\newblock \href {http://arxiv.org/abs/2402.08787} {Rethinking machine
  unlearning for large language models}.

\bibitem[{Liu et~al.(2022{\natexlab{b}})Liu, Xu, Yuan, Wang, and Li}]{liu2022}
Yi~Liu, Lei Xu, Xingliang Yuan, Cong Wang, and Bo~Li. 2022{\natexlab{b}}.
\newblock \href {https://doi.org/10.1109/INFOCOM48880.2022.9796721} {The right
  to be forgotten in federated learning: An efficient realization with rapid
  retraining}.
\newblock In \emph{IEEE INFOCOM 2022 - IEEE Conference on Computer
  Communications}, pages 1749--1758.

\bibitem[{Liu et~al.(2024{\natexlab{c}})Liu, Dou, Tan, Tian, and
  Jiang}]{liu2024safer}
Zheyuan Liu, Guangyao Dou, Zhaoxuan Tan, Yijun Tian, and Meng Jiang.
  2024{\natexlab{c}}.
\newblock \href {http://arxiv.org/abs/2402.10058} {Towards safer large language
  models through machine unlearning}.

\bibitem[{Mahabadi et~al.(2021)Mahabadi, Henderson, and
  Ruder}]{mahabadi2021compacter}
Rabeeh~Karimi Mahabadi, James Henderson, and Sebastian Ruder. 2021.
\newblock \href {http://arxiv.org/abs/2106.04647} {Compacter: Efficient
  low-rank hypercomplex adapter layers}.

\bibitem[{Maini et~al.(2024)Maini, Feng, Schwarzschild, Lipton, and
  Kolter}]{maini2024tofu}
Pratyush Maini, Zhili Feng, Avi Schwarzschild, Zachary~C. Lipton, and J.~Zico
  Kolter. 2024.
\newblock \href {http://arxiv.org/abs/2401.06121} {Tofu: A task of fictitious
  unlearning for llms}.

\bibitem[{Meng et~al.(2023)Meng, Bau, Andonian, and
  Belinkov}]{meng2023locating}
Kevin Meng, David Bau, Alex Andonian, and Yonatan Belinkov. 2023.
\newblock \href {http://arxiv.org/abs/2202.05262} {Locating and editing factual
  associations in gpt}.

\bibitem[{Mitchell et~al.(2022)Mitchell, Lin, Bosselut, Manning, and
  Finn}]{mitchell2022memorybased}
Eric Mitchell, Charles Lin, Antoine Bosselut, Christopher~D. Manning, and
  Chelsea Finn. 2022.
\newblock \href {http://arxiv.org/abs/2206.06520} {Memory-based model editing
  at scale}.

\bibitem[{Nguyen et~al.(2022)Nguyen, Huynh, Nguyen, Liew, Yin, and
  Nguyen}]{nguyen2022survey}
Thanh~Tam Nguyen, Thanh~Trung Huynh, Phi~Le Nguyen, Alan Wee-Chung Liew,
  Hongzhi Yin, and Quoc Viet~Hung Nguyen. 2022.
\newblock \href {http://arxiv.org/abs/2209.02299} {A survey of machine
  unlearning}.

\bibitem[{Patil et~al.(2023)Patil, Hase, and Bansal}]{patil2023sensitive}
Vaidehi Patil, Peter Hase, and Mohit Bansal. 2023.
\newblock \href {http://arxiv.org/abs/2309.17410} {Can sensitive information be
  deleted from llms? objectives for defending against extraction attacks}.

\bibitem[{Pawelczyk et~al.(2024)Pawelczyk, Neel, and
  Lakkaraju}]{pawelczyk2024incontext}
Martin Pawelczyk, Seth Neel, and Himabindu Lakkaraju. 2024.
\newblock \href {http://arxiv.org/abs/2310.07579} {In-context unlearning:
  Language models as few shot unlearners}.

\bibitem[{Qin and Eisner(2021)}]{qin2021learning}
Guanghui Qin and Jason Eisner. 2021.
\newblock \href {http://arxiv.org/abs/2104.06599} {Learning how to ask:
  Querying lms with mixtures of soft prompts}.

\bibitem[{Shi et~al.(2024)Shi, Ajith, Xia, Huang, Liu, Blevins, Chen, and
  Zettlemoyer}]{shi2024detecting}
Weijia Shi, Anirudh Ajith, Mengzhou Xia, Yangsibo Huang, Daogao Liu, Terra
  Blevins, Danqi Chen, and Luke Zettlemoyer. 2024.
\newblock \href {http://arxiv.org/abs/2310.16789} {Detecting pretraining data
  from large language models}.

\bibitem[{Thaker et~al.(2024)Thaker, Maurya, and Smith}]{thaker2024guardrail}
Pratiksha Thaker, Yash Maurya, and Virginia Smith. 2024.
\newblock \href {http://arxiv.org/abs/2403.03329} {Guardrail baselines for
  unlearning in llms}.

\bibitem[{Touvron et~al.(2023)Touvron, Martin, Stone, Albert, Almahairi,
  Babaei, Bashlykov, Batra, Bhargava, Bhosale, Bikel, Blecher, Ferrer, Chen,
  Cucurull, Esiobu, Fernandes, Fu, Fu, Fuller, Gao, Goswami, Goyal, Hartshorn,
  Hosseini, Hou, Inan, Kardas, Kerkez, Khabsa, Kloumann, Korenev, Koura,
  Lachaux, Lavril, Lee, Liskovich, Lu, Mao, Martinet, Mihaylov, Mishra,
  Molybog, Nie, Poulton, Reizenstein, Rungta, Saladi, Schelten, Silva, Smith,
  Subramanian, Tan, Tang, Taylor, Williams, Kuan, Xu, Yan, Zarov, Zhang, Fan,
  Kambadur, Narang, Rodriguez, Stojnic, Edunov, and
  Scialom}]{touvron2023llama2openfoundation}
Hugo Touvron, Louis Martin, Kevin Stone, Peter Albert, Amjad Almahairi, Yasmine
  Babaei, Nikolay Bashlykov, Soumya Batra, Prajjwal Bhargava, Shruti Bhosale,
  Dan Bikel, Lukas Blecher, Cristian~Canton Ferrer, Moya Chen, Guillem
  Cucurull, David Esiobu, Jude Fernandes, Jeremy Fu, Wenyin Fu, Brian Fuller,
  Cynthia Gao, Vedanuj Goswami, Naman Goyal, Anthony Hartshorn, Saghar
  Hosseini, Rui Hou, Hakan Inan, Marcin Kardas, Viktor Kerkez, Madian Khabsa,
  Isabel Kloumann, Artem Korenev, Punit~Singh Koura, Marie-Anne Lachaux,
  Thibaut Lavril, Jenya Lee, Diana Liskovich, Yinghai Lu, Yuning Mao, Xavier
  Martinet, Todor Mihaylov, Pushkar Mishra, Igor Molybog, Yixin Nie, Andrew
  Poulton, Jeremy Reizenstein, Rashi Rungta, Kalyan Saladi, Alan Schelten, Ruan
  Silva, Eric~Michael Smith, Ranjan Subramanian, Xiaoqing~Ellen Tan, Binh Tang,
  Ross Taylor, Adina Williams, Jian~Xiang Kuan, Puxin Xu, Zheng Yan, Iliyan
  Zarov, Yuchen Zhang, Angela Fan, Melanie Kambadur, Sharan Narang, Aurelien
  Rodriguez, Robert Stojnic, Sergey Edunov, and Thomas Scialom. 2023.
\newblock \href {http://arxiv.org/abs/2307.09288} {Llama 2: Open foundation and
  fine-tuned chat models}.

\bibitem[{Wang et~al.(2023)Wang, Chen, Yuan, Zeng, Wong, and Yin}]{wang2023kga}
Lingzhi Wang, Tong Chen, Wei Yuan, Xingshan Zeng, Kam-Fai Wong, and Hongzhi
  Yin. 2023.
\newblock \href {http://arxiv.org/abs/2305.06535} {Kga: A general machine
  unlearning framework based on knowledge gap alignment}.

\bibitem[{Xu et~al.(2023)Xu, Zhu, Zhang, Zhou, and Yu}]{xu2023machine}
Heng Xu, Tianqing Zhu, Lefeng Zhang, Wanlei Zhou, and Philip~S. Yu. 2023.
\newblock \href {http://arxiv.org/abs/2306.03558} {Machine unlearning: A
  survey}.

\bibitem[{Yan et~al.(2022)Yan, Li, Guo, Li, Li, and Lin}]{yan2022arcane}
Haonan Yan, Xiaoguang Li, Ziyao Guo, Hui Li, Fenghua Li, and Xiaodong Lin.
  2022.
\newblock Arcane: An efficient architecture for exact machine unlearning.
\newblock In \emph{IJCAI}, volume~6, page~19.

\bibitem[{Yao et~al.(2024)Yao, Xu, and Liu}]{yao2024llmunlearning}
Yuanshun Yao, Xiaojun Xu, and Yang Liu. 2024.
\newblock \href {http://arxiv.org/abs/2310.10683} {Large language model
  unlearning}.

\bibitem[{Zhang et~al.(2023{\natexlab{a}})Zhang, Finckenberg-Broman, Hoang,
  Pan, Xing, Staples, and Xu}]{zhang2023right}
Dawen Zhang, Pamela Finckenberg-Broman, Thong Hoang, Shidong Pan, Zhenchang
  Xing, Mark Staples, and Xiwei Xu. 2023{\natexlab{a}}.
\newblock \href {http://arxiv.org/abs/2307.03941} {Right to be forgotten in the
  era of large language models: Implications, challenges, and solutions}.

\bibitem[{Zhang et~al.(2023{\natexlab{b}})Zhang, Zhu, Zhang, Xiong, and
  Zhou}]{zhang2023}
Lefeng Zhang, Tianqing Zhu, Haibin Zhang, Ping Xiong, and Wanlei Zhou.
  2023{\natexlab{b}}.
\newblock \href {https://doi.org/10.1109/TIFS.2023.3297905} {Fedrecovery:
  Differentially private machine unlearning for federated learning frameworks}.
\newblock \emph{IEEE Transactions on Information Forensics and Security},
  18:4732--4746.

\end{thebibliography}
\bibliographystyle{acl_natbib}

\appendix
\section{Proof of Orthogonality of $v_k'$}\label{proof_orthogonality}
We offer a quick proof that $v_k'$ is orthogonal to the orthogonal components of $U$: $u_1,\cdots,u_N$. We begin with our orthogonality definition in an inner product space: 
\[u,v\text{ are orthogonal if }\braket{u,v}=0\]
Next, we consider $v_k'$ and arbitrary $u_{\ell}$:
\[v_k'=v_k-\sum_{j=1}^N \frac{\braket{v_k,u_j}}{\braket{u_j,u_j}}u_j\]
We need to show that $\braket{v_k',u_{\ell}}=0$. We proceed with the following calculation:
\begin{align*}
\langle v_k', u_{\ell} \rangle &= \left\langle v_k - \sum_{j=1}^N \frac{\langle v_k, u_j \rangle}{\langle u_j, u_j \rangle} u_j, u_{\ell} \right\rangle \\
&= \left\langle v_k, u_{\ell} \right\rangle - \left\langle \sum_{j=1}^N \frac{\langle v_k, u_j \rangle}{\langle u_j, u_j \rangle} u_j, u_{\ell} \right\rangle \\
&= \left\langle v_k, u_{\ell} \right\rangle - \sum_{j=1}^N \frac{\langle v_k, u_j \rangle}{\langle u_j, u_j \rangle} \left\langle u_j, u_{\ell} \right\rangle
\end{align*}
Since $u_1,\cdots,u_N$ are orthogonal components, we have $\braket{u_j,u_k}=0$ for $j\neq k$. This simplifies the summmation as follows:
\begin{align*}
\langle v_k', u_{\ell} \rangle &= \left\langle v_k, u_{\ell} \right\rangle - \sum_{j=1}^N \frac{\langle v_k, u_j \rangle}{\langle u_j, u_j \rangle} \left\langle u_j, u_{\ell} \right\rangle \\
&= \left\langle v_k, u_{\ell} \right\rangle - \frac{\langle v_k, u_{\ell} \rangle}{\langle u_{\ell}, u_{\ell} \rangle} \left\langle u_{\ell}, u_{\ell} \right\rangle \\
&= \langle v_k, u_{\ell} \rangle - \langle v_k, u_{\ell} \rangle \\
&= 0
\end{align*}
Thus, we have shown that $\braket{v_k',u_{\ell}}=0$ for any $u_{\ell}$, proving that $v_k'$ is orthogonal to the orthogonal components, $u_1,\cdots,u_N$, of $U$.

\end{document}